# AUTOMATED FABRIC DEFECT INSPECTION: A SURVEY OF CLASSIFIERS

Md. Tarek Habib[1], Rahat Hossain Faisal[2], M. Rokonuzzaman[3], Farruk Ahmed[4]

[1]Department of Computer Science and Engineering, Prime University, Bangladesh
[2]Department of Computer Science and Engineering, Barisal University, Bangladesh
[3]Department of Electrical and Electronic Engineering, Independent University, Bangladesh
[4]Department of Computer Science and Engineering, Independent University, Bangladesh

## ABSTRACT

*Quality control at each stage of production in textile industry has become a key factor to retaining the existence in the highly competitive global market. Problems of manual fabric defect inspection are lack of accuracy and high time consumption, where early and accurate fabric defect detection is a significant phase of quality control. Computer vision based, i.e. automated fabric defect inspection systems are thought by many researchers of different countries to be very useful to resolve these problems. There are two major challenges to be resolved to attain a successful automated fabric defect inspection system. They are defect detection and defect classification. In this work, we discuss different techniques used for automated fabric defect classification, then show a survey of classifiers used in automated fabric defect inspection systems, and finally, compare these classifiers by using performance metrics. This work is expected to be very useful for the researchers in the area of automated fabric defect inspection to understand and evaluate the many potential options in this field.*

## 1. INTRODUCTION

Quality assurance of product is considered as one of the most important focuses in the industrial production. So is textile industry too. Textile product quality is seriously degraded by defects. Failure to early defect detection costs time, money and consumer satisfaction. So, early and accurate fabric defect detection is an important phase of quality control. Manual inspection is time consuming and the level of accuracy is not satisfactory enough to meet the present demand of the highly competitive international market. Hence, expected quality cannot be maintained with manual inspection. Automated, i.e. computer vision based fabric defect inspection system is the solution to the problems caused by manual inspection. Automated fabric defect inspection system has been attracting extensive attention of the researchers of many countries for years. Automated fabric defect inspection system mainly involves two challenging problems, namely defect detection and defect classification.

Even though many classifiers have already been used, none of the classification techniques is that much effective rather they are relative. In fact, deployment of these techniques varies according to the environment. In this paper, we describe different defect classification techniques. We hope that the paper would be useful for researchers and developers to choose the best one among various options of defect classification techniques while working on an automated fabric defect inspection system.

## 2. LITERATURE REVIEW

Some efforts have been made for the survey of automated, i.e. computer vision based inspection systems [1, 2], where not a large number of attempts have been made for survey of automated, i.e. computer vision based fabric defect inspection systems [3-6]. All of them have concentrated on defect detection, where none of them has concentrated on defect classification. Kumar [3] has used about 150 references for his comprehensive survey of automated fabric defect inspection systems. His survey has been limited to defect detection methodologies only. Likewise, the surveys of Mahajan et al. [4], Ngan et al. [5] and Shanbhag et al. [6] have been done on defect detection techniques only. Moreover, their information collection and organization have not been well enough.

## 3. FABRIC DEFECTS

The presence of fabric defect makes the final garment product faulty. It has been reported in [7] that fabric defects cause about 85% of the defects detected in the garment industry. Fabric prices decrease by 45%–65% for second or off quality goods [8]. It is therefore of great importance that these defects are detected, recognized and prevented from reoccurring. There are a large number of types of fabric defects. Among them, color yarn, missing yarn, hole, slub, crease mark and spot are the notable ones, which are shown in Figure 1. The quality of textile products important phase of quality improvement. An automated fabric defect inspection system increase the product quality, which poses improved productivity to conform to customer needs and to lessen the costs associated with quality degradation. In automated fabric defect inspection, as the fabric is produced, inspection is automatically done. This process is continued by detecting the defect in fabric and then by classifying the fabric based on some features of the defect. It has been reported in [4] that the investment in the automated fabric defect inspection system is beneficial taking decrease in labor cost and associated advantages into account.

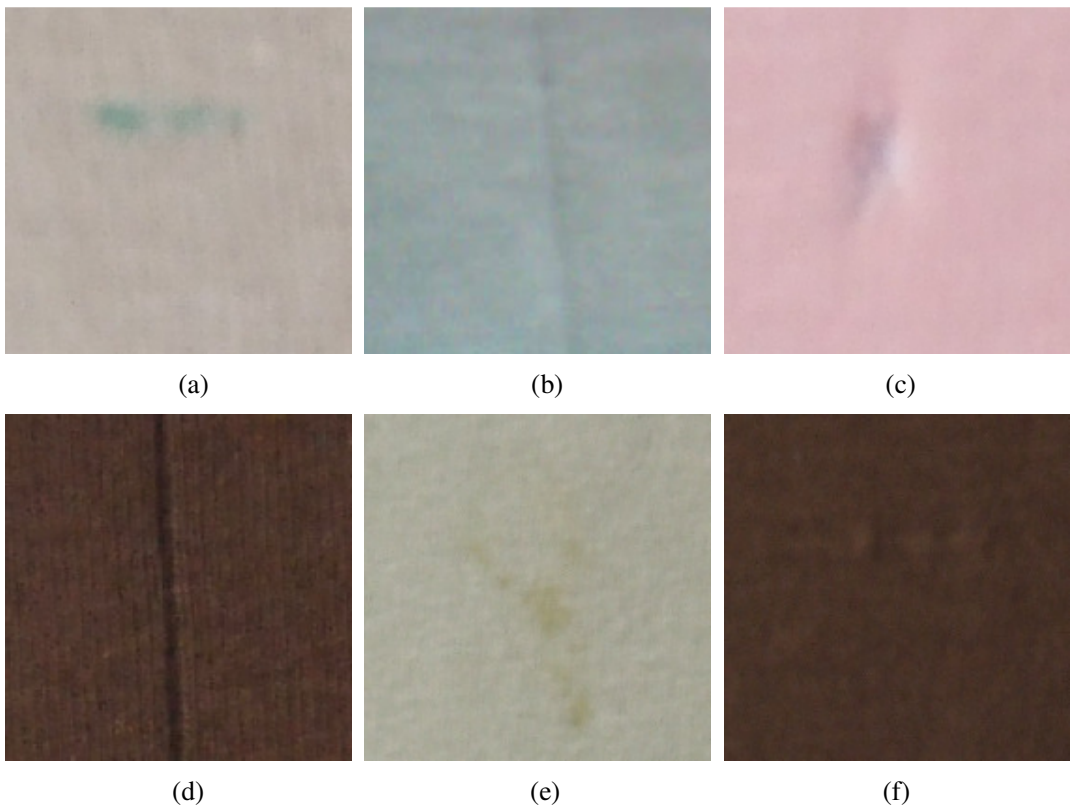

Figure 1. Some notable types of defects occurred in fabrics. (a) Color yarn. (b) Crease mark. (c) Hole. (d) Missing yarn. (e) Spot. (f) Slub.

## 4. FABRIC DEFECT INSPECTION PROCESS

The development of an automated, i.e. computer vision based system for fabric defect inspection involves several steps as shown in Figure 2. Each step has effects on the performances of the subsequent steps. Weak design and implementation of a step make the subsequent steps complicated, which results in harder development of the system. So each step has a lot of importance in the development of a fully automated system for textile industry. An automated fabric inspection system is comprised with the following steps shown in Figure 2.

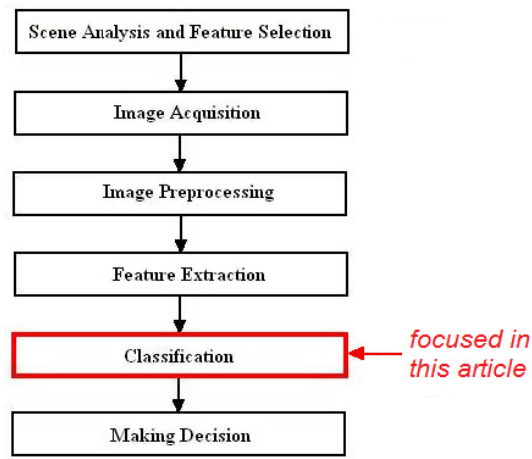

Figure 2. Different stages of an automated fabric defect inspection system.

In this work, we mainly focus on only one step of the development of computer vision based fabric defect inspection system. That is the classification stage as shown in Figure 2.

A number of attempts have been made for computer vision based fabric defect inspection [9-31]. Only few of them have focused on classification, where the majority have concentrated on defect detection. Mainly three defect-detection techniques [3, 11], namely statistical, spectral and model-based, have been deployed. A number of techniques have been used for classification. Among them, artificial neural network (ANN), support vector machine (SVM), clustering and statistical inference are notable.

## 5. PERFORMANCE METRICS OF CLASSIFIERS

Different performance metrics can be considered to measure the performance of a classifier. Classification accuracy, model complexity and training time are considered as three most important performance metrics for an automated fabric defect inspection system.

## 6. A SURVEY OF CLASSIFIERS

Different techniques have been applied in classification. Among them - ANNs, support vector machines (SVMs), clustering and statistical inference are the prominent ones.

### 6.1. Statistical Inference

In [21] and [22], statistical inference is used for classification. Cohen et al. [21] have chosen statistical test for classification. They have used likelihood-ratio test as statistical test in order to implement binary classification, i.e. categorization on only one property, defective or defect-free. Campbell et al. [22] have used hypothesis testing for classification. They also have implemented classification on only defective or defect-free classes. Categorization of defect-free and defective fabrics only is not the ultimate goal of fabric defect classification.

### 6.2. Support Vector Machine

Murino et al. [13] have used SVMs for classification. They have worked on spatial domain. They have used features of three types. The features have been extracted from gray-scale histogram, shape of defect and co-occurrence matrix. The feature extraction process has been supposed to be complex due to some of the features. The basic SVM scheme is designed for binary classification problem. They have used SVMs to solve multiclass problem, i.e. distinct categorization of defects. They have implemented SVMs with 1-vs-1 binary decision tree scheme in order to solve the multiclass problem. Two data sets, i.e. sets of images have been separately used in their entire work. One set contained 11 types of defects and the other set contained 9 types of defects.

### 6.3. Clustering

Campbell et al. [20] have applied model-based clustering for defect classification. Clustering is not a good choice for real-time systems like computer vision based fabric inspection systems.

### 6.4. Artificial Neural Network

ANNs have been deployed as classifiers in a number of articles. Various learning algorithms have been applied to train the ANNs.

Habib and Rokonuzzaman [9] have worked with four types of defects and used counterpropagation neural network (CPN) to classify the defects. They focused on feature selection instead of paying attention to the CPN model. They have not done thorough investigation on the applicability of CPN model in the automated fabric defect inspection domain. Backpropagation learning algorithm has been used in [11], [14], [17], [18] and [26]. Saeidi et al. [11] have trained their ANN by backpropagation algorithm so as to deal with multiclass problem, i.e. categorizing defects distinctly. They have done on-line implementation after performing off-line experiments. In both cases, they have used 6 types of defects. Their work is on frequency domain. Karayiannis et al. [14] have used an ANN trained by backpropagation algorithm in order to solve multiclass problem. They have used 7 types of defects. They have used statistical texture features. An ANN trained by backpropagation algorithm has been deployed by Kuo and Lee [17] in order to deal with multiclass problem. 4 types of defects have been used by them. Maximum length, maximum width and gray level of defects have been considered as features. So the feature number has become too small. They have found good classification accuracy because the sample size was also small. There can be the case that their approach will not successfully classify

International Journal in Foundations of Computer Science & Technology (IJFCST), Vol.4, No.1, January 2014

defects due to this small number of features when the sample size tends to be much large. Mitropulos et al. [18] have trained their ANN by backpropagation algorithm so as to deal with multiclass problem. 7 types of defects have been used by them. Although they have used first and second order statistical features, the feature number has been small. Moreover their sample size was small too. It seems that their approach performed good due to the small sample size. There can be the case that their approach will work poorly for this small number of features when the sample size tends to be much large. Habib and Rokonuzzaman [26] have used fully connected feedforward ANN to deal with multiclass problem. They have trained their ANN by backpropagation algorithm. They have used 4 types of defects. They have used features involving defect size and shape.

Resilient backpropagation learning algorithm has been used in [12] and [31]. Islam et al. [12] have designed their ANN trained by resilient backpropagation algorithm in order to deal with multiclass problem. They have worked with more than 2 types of defects. That means they have considered two types of defects as two major types and all other types of defects as a single major type. The area, number of parts and sharp factor of defect have been used by them as features and therefore the feature number has become too small. Moreover, they have justified the features very little. They got some success because the sample size was small. There is a great chance that their approach will poorly classify defects due to this small number of features when the sample size tends to be much large. Islam et al. [31] have also employed an ANN trained by resilient backpropagation algorithm in order to address multiclass problem. More than 2 types of defects have been used by them. That means two types of defects have been considered as two major types and all other types of defects have been considered as a single major type. They have used three features, namely the area of faulty portion, number of objects and shape factor. So the feature number has become too small. In spite of the too small feature number, their approach worked. It seems that the small sample size was the reason behind this. It is supposed to happen that their approach will not achieve the result as per expectation due to this small feature number when the sample size goes much large.

Learning vector quantization (LVQ) algorithm was applied by Shady et al. [19] so as to train their ANNs. Their ANNs have been implemented in order to handle multiclass problem. They have used 6 types of defects. Separately work has taken place in defect detection process for spatial and frequency domains. That means statistical technique and spectral technique, i.e. Fourier transform, have separately been deployed for detecting defects. In case of statistical technique, a grid measuring scheme has been used for calculating the row and column vectors of images. Statistical features, e.g. mean, median etc are extracted from the row and column vectors. Kumar [15] has used two ANNs separately. The first one trained by backpropagation algorithm has been designed for binary classification, i.e. categorization only on defective or defect-free. It has been shown that the inspection system with this ANN is not effective in terms of cost. So he has further used linear ANN. He has applied least mean square error (LMS) algorithm so as to train the network. The inspection system with this ANN is cost-effective, but it cannot deal with multiclass problem. Inability to deal with multiclass problem, i.e. binary-classification ability, does not fulfil the ultimate need of fabric defect classification. Karras et al. [16] have also separately used two ANNs. One ANN has been trained by backpropagation algorithm. The other ANN used by them is Kohonen's Self-Organizing Feature Maps (SOFM). First and second order statistical-texture features have been used for both ANNs. Both of the ANNs have been designed to solve binary classification problem, i.e. categorization of only defective and defect-free, which does not fulfil the ultimate need of fabric defect classification. Table 1 summarizes the discussion in this paragraph.

## 7. COMPARISON OF CLASSIFIERS

The articles discussed in the previous section are typical examples of effort for computer vision based fabric defect inspection. Considering these articles as a sample, we make some comparative analysis as shown in Fig. 3. A little more effort has been made to deal with the multiclass problem, i.e. categorizing defects distinctly, than to implement binary classification, i.e. categorization of only defective and defect-free classes. This is shown in Fig. 3(a). We see from Fig. 3(b) that ANNs have been chosen for classification much more than any other methods.

Table 1. Summary of the articles discussed.

| Reference | Capability of Classifying Defects Distinctly | Classifier | Learning Algorithm |
|---|---|---|---|
| [9] | √ | ANN | Counterpropagation |
| [11] | √ | ANN | Backpropagation |
| [12] | √ | ANN | Resilient backpropagation |
| [13] | √ | SVMs | NA |
| [14] | √ | ANN | Backpropagation |
| [15] | × | ANN | Backpropagation |
| [15] | × | ANN | Least mean square error (LMS) |
| [16] | × | ANN | Backpropagation |
| [16] | × | ANN | Kohonen's Self-Organizing Feature Maps (SOFM) |
| [17] | √ | ANN | Backpropagation |
| [18] | √ | ANN | Backpropagation |
| [19] | √ | ANN | Learning vector quantization (LVQ) |
| [20] | √ | Model-based clustering | NA |
| [21] | × | Statistical test | NA |
| [22] | × | Statistical test | NA |
| [26] | √ | ANN | Backpropagation |
| [31] | √ | ANN | Resilient backpropagation |

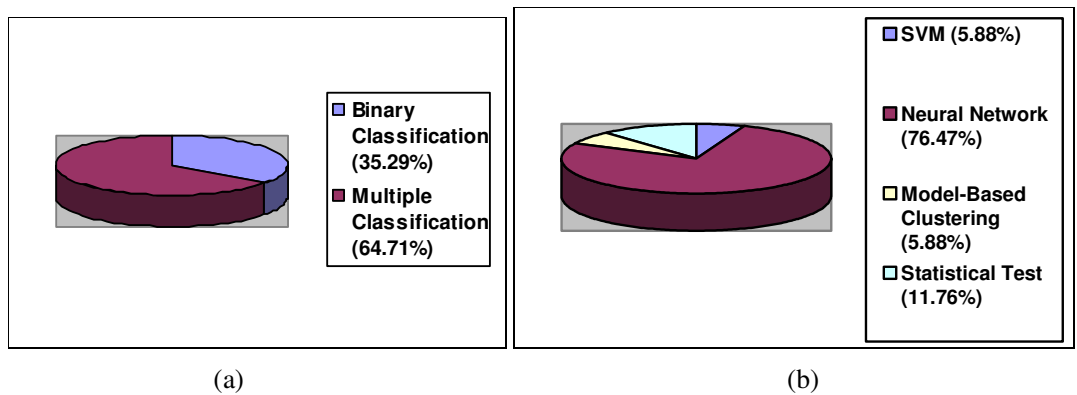

(a)          (b)

Figure 3. Comparative analysis of the articles discussed. (a) Dealing with classification problem. (b) Using classification methods.

## 8. COMPARATIVE ANALYSIS OF ARTIFICIAL NEURAL NETWORKS

We have found from the comparative discussion of the articles in the previous section that ANNs have been chosen for classification much more than any other methods. Considering these articles as a sample, we make one more comparative analysis as shown in Fig. 4. We see from Fig. 4 that backpropagation is the mostly used learning algorithm when ANNs are used as classifier. Finally, we compare the performance of the ANN models in terms of three performance matrices – accuracy, model complexity and training time as shown in Table 2.

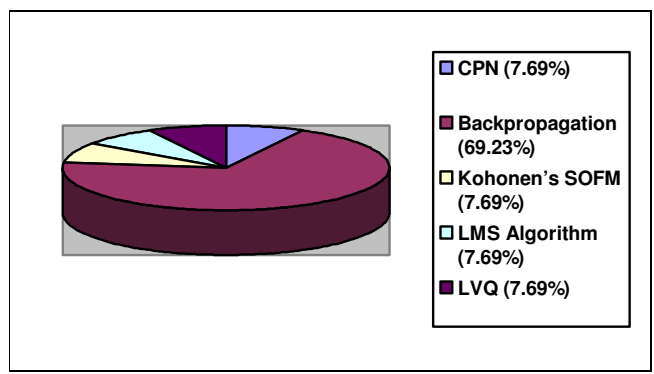

Figure 4. Comparison of the learning algorithms used for ANNs in the articles discussed.

Table 2. Comparison of ANN based classification models.

| Article | Fabric Type | Number of Input Sites | Number of Classes | Performance Metrics | | |
|---|---|---|---|---|---|---|
| | | | | Training Time (Number of Elapsed Cycle) | Model Complexity (Number of Computing Units) | Accuracy |
| [9] | Knitted fabric | 4 | 6 | 6 | 4-7-6 | 98.97% |
| [11] | Knitted fabric | 15 | 7 | 7350 | 15-8-7 | 78.4% |
| | | NM[1] | NM | NM | NM | 96.57% |
| [12] | NM | 3 | 4 | NM | 3-40-4-4 | 77% |
| [14] | Web textile fabric | 13 | 8 | NM | 13-5-8 | 94% |
| [18] | Web textile fabric | 4 | 8 | NM | 4-5-8 | 91% |
| [19] | Knitted fabric | 7 | 7 | NM | 7-7 | 90.21% |
| | | 6 | 7 | NM | 6-7 | 91.9% |
| [26] | Knitted fabric | 4 | 6 | 137043 | 4-12-6 | 100% |
| [31] | NM | 3 | 4 | NM | 3-44-4 | 76.5% |

[1]NM: Not Mentioned

## 9. CONCLUSION

In this paper, major defect classification techniques for automated fabric defect inspection system have been considered initially. After that defect classifiers of automated fabric defect inspection systems have been surveyed. Finally, we have compared these classifiers based on three prominent performance metrics – accuracy, model complexity and training time. We hope that this paper would be useful for every newcomer to the area of automated fabric defect inspection.

## REFERENCES

[1] X. Xie, "A Review of Recent Advances in Surface Defect Detection using Texture analysis Techniques," *Electronic Letters on Computer Vision and Image Analysis*, vol. 7, no. 3, pp. 1-22, 2008.

[2] T. S. Newman and A. K. Jain, "A Survey of Automated Visual Inspection," *Computer Vision and Image Understanding*, vol. 61, no. 2, pp. 231-262, March 1995.

[3] A. Kumar, "Computer-Vision-Based Fabric Defect Detection: A Survey," *IEEE Transactions on Industrial Electronics*, vol. 55, no. 1, pp. 348-363, January 2008.

[4] P. M. Mahajan, S. R. Kolhe, and P. M. Patil, "A review of automatic fabric defect detection techniques," *Advances in Computational Research*, vol. 1, no. 2, pp. 18-29, 2009.

[5] H. Y. T. Ngan, G. K. H. Pang, and N. H. C. Yung, "Automated fabric defect detection-A review," *Image and Vision Computing*, vol. 29, no. 7, pp. 442-458, 2011.

[6] P. M. Shanbhag, M. P. Deshmukh, and S. R. Suralkar, "Overview: Methods of Automatic Fabric Defect Detection," *Global Journal of Engineering, Design & Technology*, vol. 1, no. 2, pp. 42-46, 2012.

[7] P. Sengottuvelan, A. Wahi, and A. Shanmugam, "Automatic Fault Analysis of Textile fabric Using Imaging Systems," *Research Journal of Applied Sciences*, vol. 3, no. 1, pp. 26-31, 2008.

[8] K. Srinivasan, P. H. Dastor, P. Radhakrishnaihan, and S. Jayaraman, "FDAS: A knowledge-based frame detection work for analysis of defects in woven textile structures," *Journal of Textile Institute*, vol. 83, no. 3, pp. 431-447, 1992.

[9] M. T. Habib and M. Rokonuzzaman, "A Set of Geometric Features for Neural Network-Based Textile Defect Classification," ISRN Artificial Intelligence, Volume 2012, Article ID 643473, 16 pages, 2012.

[10] R. Stojanovic, P. Mitropulos, C. Koulamas, Y.A. Karayiannis, S. Koubias, and G. Papadopoulos, "Real-time Vision based System for Textile Fabric Inspection," *Real-Time Imaging*, vol. 7, no. 6, pp. 507-518, 2001.

[11] R. G. Saeidi, M. Latifi, S. S. Najar, and A. Ghazi Saeidi, "Computer Vision-Aided Fabric Inspection System for On-Circular Knitting Machine," *Textile Research Journal*, vol. 75, no. 6, pp. 492-497, 2005.

[12] M. A. Islam, S. Akhter, and T. E. Mursalin, "Automated Textile Defect Recognition System using Computer Vision and Artificial Neural Networks," *Proceedings World Academy of Science, Engineering and Technology*, vol. 13, pp. 1-7, May 2006.

[13] V. Murino, M. Bicego, and I. A. Rossi, "Statistical Classification of Raw Textile Defects," icpr, pp. 311-314, 17[th] *International Conference on Pattern Recognition (ICPR'04)*, vol. 4, 2004.

[14] Y. A. Karayiannis, R. Stojanovic, P. Mitropoulos, C. Koulamas, T. Stouraitis, S. Koubias, and G. Papadopoulos, "Defect Detection and Classification on Web Textile Fabric Using Multiresolution Decomposition and Neural Networks," *Proceedings on the 6th IEEE International Conference on Electronics, Circuits and Systems*, Pafos, Cyprus, September 1999, pp. 765-768.

[15] A. Kumar, "Neural Network based detection of local textile defects," *Pattern Recognition*, vol. 36, pp. 1645-1659, 2003.

[16] D. A. Karras, S. A. Karkanis, and B. G. Mertzios, "Supervised and Unsupervised Neural Network Methods applied to Textile Quality Control based on Improved Wavelet Feature Extraction Techniques," *International Journal on Computer Mathematics*, vol. 67, pp. 169-181, 1998.

[17] C.-F. J. Kuo and C.-J. Lee, "A Back-Propagation Neural Network for Recognizing Fabric Defects," *Textile Research Journal*, vol. 73, no. 2, pp. 147-151, 2003.

[18] P. Mitropoulos, C. Koulamas, R. Stojanovic, S. Koubias, G. Papadopoulos, and G. Karayiannis, "Real-Time Vision System for Defect Detection and Neural Classification of Web Textile Fabric," *Proceedings SPIE*, vol. 3652, San Jose, California, pp. 59-69, January 1999.
[19] E. Shady, Y. Gowayed, M. Abouiiana, S. Youssef, and C. Pastore, "Detection and Classification of Defects in Knitted Fabric Structures," *Textile Research Journal*, vol. 76, No. 4, pp. 295-300, 2006.
[20] J. G. Campbell, C. Fraley, D. Stanford, F. Murtagh, and A. E. Raftery, "Model-Based Methods for Textile Fault Detection," *International Journal of Imaging Systems and Technology*, vol. 10 Issue 4, pp. 339-346, Jul 1999.
[21] F. S. Cohen, Z. Fan, and S. Attali, "Automated Inspection of Textile Fabrics Using Textural Models," *IEEE Trans. Pattern Anal. Mach. Intell.*, vol. 8, no. 13, pp. 803-808, Aug. 1991.
[22] J. G. Campbell, A. A. Hashim, T. M. McGinnity, and T. F. Lunney. "Flaw Detection in Woven Textiles by Neural Network," in *Fifth Irish Neural Networks Conference*, St. Patrick's College, Maynooth, pp. 92-99, Sept. 1995.
[23] K. L. Mak, P. Peng, and H. Y. K. Lau, "A Real-Time Computer Vision System for Detecting Defects in Textile Fabrics," *IEEE International Conference on Industrial Technology*, Hong Kong, China, 14-17, pp. 469-474, Dec. 2005.
[24] A. Baykut, A. Atalay, A. Erçil, and M. Güler, "Real-Time Defect Inspection of Textured Surfaces," *Real-Time Imaging*, vol. 6, no. 1, pp. 17-27, Feb. 2000.
[25] F. S. Cohen and Z. Fan, "Rotation and Scale Invariant Texture Classification," in *Proc. IEEE Conf. Robot. Autom.*, vol. 3, pp. 1394-1399, April 1988..
[26] M. A. Islam, S. Akhter, T. E. Mursalin, and M. A. Amin, "A Suitable Neural Network to Detect Textile Defects," *Neural Information Processing, SpringerLink,* vol. 4233, pp. 430-438, October 2006.
[27] A. Abouelela, H. M. Abbas, H. Eldeeb, A. A. Wahdan, and S. M. Nassar, "Automated Vision System for Localizing Structural Defects in Textile Fabrics," *Pattern Recognition Letters*, vol. 26, Issue 10, pp. 1435-1443, July 2005.
[28] W. Jasper, J. Joines, and J. Brenzovich, "Fabric Defect Detection Using a Genetic Algorithm Tuned Wavelet Filter," *Journal of the Textile Institute*, vol. 96, Issue 1, pp. 43-54, January 2005.
[29] Y. Shu and Z. Tan, "Fabric Defects Automatic Detection Using Gabor Filters," *World Congress on Intelligent Control and Automation (WCICA 2004),* Hangzhou, China, vol. 4, pp. 3378-3380, June 2004.
[30] M. Salahuddin and M. Rokonuzzaman, "Adaptive Segmentation of Knit Fabric Images for Automated Defect Detection in Semi-structured Environments," *Proceedings of the 8th ICCIT*, pp. 255-260, 2005.
[31] M. T. Habib and M. Rokonuzzaman, "Distinguishing Feature Selection for Fabric Defect Classification Using Neural Network", Journal of Multimedia, vol. 6, no. 5, October 2011.

**Author**

**Md. Tarek Habib** received his B.Sc. degree in Computer Science and M.S. degree in Computer Science and Engineering from BRAC University and North South University in 2006 and 2009 respectively. He is an Assistant Professor in the Department of Computer Science and Engineering of Prime University, Bangladesh. His research interest is in artificial intelligence, especially neural networks and computer vision.

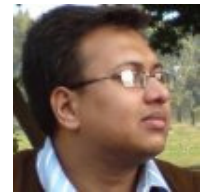